# Semantically enriched spatial modelling of industrial indoor environments enabling location-based services

Arne Wendt, Michael Brand
and Thorsten Schüppstuhl

Institute of Aircraft Production Technology, Hamburg University of Technology
Denickestraße 17, 21073 Hamburg, Germany
Email: {arne.wendt, michael.brand, schueppstuhl}@tuhh.de

**Abstract.** This paper presents a concept for a software system called RAIL representing industrial indoor environments in a dynamic spatial model, aimed at easing development and provision of location-based services. RAIL integrates data from different sensor modalities and additional contextual information through a unified interface. Approaches to environmental modelling from other domains are reviewed and analyzed for their suitability regarding the requirements for our target domains; intralogistics and production. Subsequently a novel way of modelling data representing indoor space, and an architecture for the software system are proposed.

## 1 Introduction

The location-based services (LBS) available on GPS-enabled mobile phones provide value to users by assisting them based on their location and other contextual information. Likewise, one can imagine LBS could prove beneficial to workers and mobile robots in industrial indoor environments as well. Enabling utilization of the wide array of (positional) sensors already present in autonomous systems used today, holds a lot of potential for providing these LBS.

In this paper a dynamic environmental model, named RAIL, is proposed. The model is to be implemented as a software system, aiming to standardize the dynamic representation of industrial indoor environments in an effort to enable a range of LBS.

RAIL features the following constitutive properties:

- Object-oriented environmental model
- Spatial information is represented by relative 6D transformations between objects
- Integration of information from different sensor modalities in a common model

- Implementation of the dynamic map as a software system providing unified interfaces each for feeding the map with data and querying data from the map
- Conceptually supporting any kind of data to enable near arbitrary LBS

## 2 State of the Art and Related Work

With the goal of providing LBS for industrial indoor environments, it seems reasonable to look at what is established in the outdoor world today. Outdoor LBS are ubiquitous in today's society, accessible through commercial products, with Google Maps being at the forefront here. Outdoor LBS can serve as an example to get a grasp of the possibilities of LBS in general: Basic outdoor LBS include navigation services, services providing geospatial range querys paired with additionally available data (POIs of a certain type in range, *e.g. "Find open supermarket within 1000 m"*) and user notifications based on geofencing (advertising [1], disaster management [2, 3, 4]). For a broader classification, Barnes [5] divides outdoor LBS into four domains: Safety, navigation and tracking, transactions and information. A similar classification is employed in [1]. It comes out clearly that there is a spectrum ranging from very basic LBS, merely passing on location information, up to LBS extensively combining spatial with other information. Like with outdoor LBS, development of indoor LBS and the assessment of their potential focus mainly on the consumer market. Most commercially available systems target the services navigation and geofencing for user notification (mainly advertising). Examples of commercially available solutions are: infsoft [6], IndoorLBS [7], IndoorAtlas [8], estimote [9], MazeMap [10] and micello [11]; which seem to be predominantly employed in shopping malls. Other applications include disaster management, human trajectory identification and facility management [12]. Taking a look at LBS targeted at industrial applications there are commercially available systems providing the services positioning and identification as well as geofencing for indoor production environments. Two examples of this are Ubisense SmartFactory [13] and Quantitec IntraNav [14]. Both solutions are extensible via their SDK, but remain proprietary solutions.

With the aim of providing LBS, an underlying spatial model of the indoor environment suited for the targeted service-application is needed. Modelling of indoor environments is an active field of research where two predominant use-cases can be described [12]: First, providing context sensitive information and aid to people acting in indoor space. Second, gathering information about the building itself for maintenance applications. In [15] two main types of modelling indoor space are inferred: Namely, geometric and symbolic spatial models. Geometric models are further divided into cell-based and boundary-based representations. With cell-based models, the physical space is dissected into cells, where every cell gets labeled according to its function (e.g. free space, walls, doors). The cell size is chosen to fit the needed accuracy of the application. In boundary-based representations, only the boundaries of a space (e.g. walls, obstacles) are modelled with geometrical primitives. This yields a very compact map, but no semantic information is included. Symbolic models focus on representing the environ-ment and objects of interest in the form of topological-based structures, graphs or hierarchies. That way, relations such as connectivity, adjacency, and containment can be modelled, thus representing the logical structure of a building's components. Object-oriented symbolic models put a special focus on representing individual objects including semantics. This stands in contrast to only representing building components (i.e. rooms, floors). Symbolic models do not necessarily contain any geometric information. Reviewing literature on the topic of modelling indoor environments we identify the focus as human navigation and assistance in static environments.

Another field that has to be considered when surveying environmental (especially spatial) representations is that of mobile robotics. The maps used by mobile robots focus on enabling specifically localization and navigation. However, they usually don't feature any semantic information nor notion of objects. As well in the field of robotics, the ROS-tf-library [16] is mentionable. Though not necessarily targeted at representing indoor space, it aims at representing (tempo-) spatial relationships in a symbolic form as a tree of relative transformations between coordinate frames. The tf-library does not support modelling anything besides transformations, thus not lending itself to model any environment as a whole by itself – lacking capabilities of modelling geometry and semantics.

## 3 Our approach

The main goal of RAIL is to simplify development and provision of LBS in industrial indoor environments by providing open infrastructure to be able to model, acquire and share the needed information and data. The open nature of the system is a key element in setting it apart from commercially available solutions, as we believe that removing commercial interest on the platform side will foster the emergence of a more innovative and diverse range of services and the integration of a more diverse range of data sources. For that purpose, in the following sections, a suitable environmental modelling approach, and a concept for the software system implementing this model are proposed. It has to be stressed that RAIL does not describe a data format for storing a persistent spatial model suitable for data exchange - e.g. CityGML [18], GeoJSON [19] - but is a (distributed) software system managing the data representing the current environmental state. This section starts with our take on LBS and the target scope of RAIL regarding LBS-consumers, environment and use-cases. This is followed by an evaluation of existing modelling approaches (outlined in section 2) in terms of the applications requirements. The successive subsections will cover the data modelling and software system's design.

LBS shall provide actors with required information, enabling them to fulfill their assigned tasks. These actors are explicitly not restricted to be human, but include autonomous (robotic) systems as well. Focusing on industrial applications, especially production, intralogistics and MRO (Maintenance, Repair and Operations), the tasks to fulfill will likely require interaction with and manipulation of objects in the environment – assisting interaction with the physical environment being one of the problems in modelling, as identified in [17]. The environment to be acted in is assumed to be a factory floor, including machinery, storage racks, assembly areas, etc. Combining the assumption that tasks involve object interaction/manipulation with the integration of autonomous systems, the need for not only positional information, but for full 6D-poses can be deduced - which in turn becomes a key requirement for the proposed model. To represent factory floors (some hundred meters across) and at the same time enable handling operations in production (which requires (sub-) millimeter resolution), simultaneously a high resolution and wide spatial range are required to be representable by the model.

Checking these requirements against geometrical models, it becomes evident that 6D-transformations of objects can't be stored in the described geometrical models which renders them unsuitable. On top of that, the conflicting requirements of a wide spatial range and fine resolution would likely result in a poor performance with a purely geometrical representation. When checking the requirements against symbolic models, an object-oriented approach seems viable for representing an industrial environment and the individual objects of interest. Geometric information can be included in the form of 6D-transformations connected with each object in the model. Different from the classification in [15], we see no use for a hierarchical structure of an object-oriented model for factory-floor modelling. As opposed to modelling e.g. office buildings, factory-floors usually lack structural elements like floors and rooms, thus not benefiting from hierarchical modelling on basis of containment.

Looking at spatial modelling techniques in mobile robotics, the environmental models usually have no notion of individual objects and lack methods to model semantic information, both of which are key for enabling advanced LBS that aid context-sensitive interaction with objects. Thus, these modelling approaches are unsuitable in the present case.

The requirements to be satisfied by RAIL and the chosen modelling, arising from the targeted applications and industrial environment shall be as follows:

- Provide a dynamic model of the indoor environment
- Represent relative spatial relationships in the form of 6D-transformations
- Integrate (sensory) data from different domains
- Allow deposition of CAD-models, images, feature descriptors, …
- Robust against shutdown/failure of software components
- Robust against hardware shutdown/failure and partial infrastructure outage
- Consider scalability for data providers, data consumers and amount of data in the model

## 3.1 Data Modelling

A major problem in designing the data modelling is the demand to enable a as broad as possible range of LBS. The challenge here is that the concrete LBS to be supported are unknown at this point. Since all modelled data serves the goal of enabling LBS, it is unknown what exactly has to be modelled. Thus we will limit RAIL here to only operate on spatial relationships and geometric primitives to enable basic operations such as *spatial-range-queries*, but won't otherwise interpret the supplied data - which shall rather be done by the LBS or their consumers.
With the goal of assisting in interaction with and manipulation of objects in the environment by providing LBS, there is a strong focus on the actual objects making up the environment. We need to be able to either identify these objects or at least categorize them to be of any use for the intended application, as only location and identity of objects combined allow reasoning about the environment on a semantical level. Hence, we propose an object-oriented modelling approach, explicitly modelling all necessary individual objects with their specific properties, thus adding value to the LBS present.

In this context an object shall be an abstract concept. That is, an object can be purely logical and does not necessarily need a physical representation. Object shall be a logical clustering entity comprised of arbitrary attributes and assigned values. This approach lacks a higher level (hierarchical) ordering principle, but by allowing arbitrary attributes to be assigned to objects, we preserve the option to model logical relationships (containment, adjacency, connectivity, etc.) between objects. Conceptually, the location of an object shall be an attribute of the object.

Regarding the implementation of a dynamic object-oriented spatial model, we see the need to separate the spatial information from the remaining attributes of the modelled entities. The first reason for this is the requirement to model relative spatial relationships. Modelling relative spatial relationships can help in reducing uncertainty and increasing accuracy of transformations retrieved from RAIL by lowering the number of individual transformations (each with their associated uncertainty and resolution) required to compute the actually desired transformation. As noted earlier, the spatial relationships/transformations between objects shall all be in 6D, each carrying additional information about uncertainty and resolution. These relationships effectively construct a graph. The nodes of this graph are the coordinate frames of entities other entities' positions have been determined in, thus representing the modelled objects. The edges represent the transformations between these frames. For computations on the graph, the weight assigned to the edges shall be the uncertainty and resolution of the transformations, allowing to retrieve the transformation with the highest resolution and lowest uncertainty between two frames/objects. As we can see here, querying for transformations (i.e. positional data) requires computation apart from that required for performing database searches. This stands in contrast to queries regarding all other information present about modelled entities. This of course bases upon the fact, that we do not directly interpret any other than spatial information, but rather only relay it. The second reason is that we want to enable spatial range queries which depend on the ability to calculate transformations while taking into account the dimensions of entities as well.

In summary, RAIL will consist of a type of document store, holding arbitrary information about the modelled entities, and a graph of all known transformations between these entities. The document store will be holding the entities as containers for the semantics associated with the elements making up the object-oriented model. The graph containing the transformations between coordinate frames of known entities adds the spatial component to the model. All spatial data is held in a symbolic form, thus not limiting accuracy[1] by design.

### 3.2 Systems Design

This section will begin with a presentation of the overall structure and the interfaces to RAIL. This is followed by an overview of the proposed software system's design, targeted at satisfying the requirements of building a dynamic environmental model for industrial applications.

As noted at the beginning of the last section, RAIL has to be tailored in a way enabling it to serve as the data basis for a diverse range of LBS, with the concrete LBS to be supported being unknown at this point in time. The requirement arising from this is that we need to be able to feed arbitrary data into RAIL, as well as be able to retrieve all data present. This requirement allows to reduce the interfaces to two general interfaces; one for data providers and one for consumers. This greatly reduces the complexity of the interfaces, as they both serve a distinct purpose.

With RAIL not only targeted at modelling industrial environments, but to be used in industrial applications, we have to satisfy the demand for high failure tolerance and robustness by design. Further we will try to enable horizontal scaling regarding most parameters influencing performance to cope with increased data volumes and computational load. We will require all interaction with RAIL to be on TCP/IP and UDP, and explicitly do not target operation on a field-bus-level. For the conceptual design of the interfaces we will further make the following assumptions: Data providers are likely to be on wireless networks with low reliability, and consumers requiring high frequency updates, using the publish/subscribe-interface, are on highly reliable connections.

To account for these characteristics of the data providers/sensors, we want the according interface to be state- and session-less on the application-layer to allow connection drop outs without requiring any time consuming re-establishment of a session. Simultaneously enabling flexible data routing and eased hand- and failover of software components. For data consumers, the assumed high reliability connection allows to spare efforts on making the interface state- or session-less. As we are developing a centralized model, there will be no explicit many-to-many messaging from a data providers point of view, meaning we do not need publish/subscribe-messaging on the systems input and data providers shall "push" their data. Therefore, and with the intention of building state- and session-less interfaces, RAIL will not allow data providers to stream their data, but limit all communication on the ingoing side to message-based communications. On the consumer side we will provide a request/response-interface for queries to the model as well as a publish/subscribe-like-interface. The interface provides a "change-feed" to the consumer continuously providing it with updates to the query it posed initially.

[1] Except for limits posed by computer architectures.

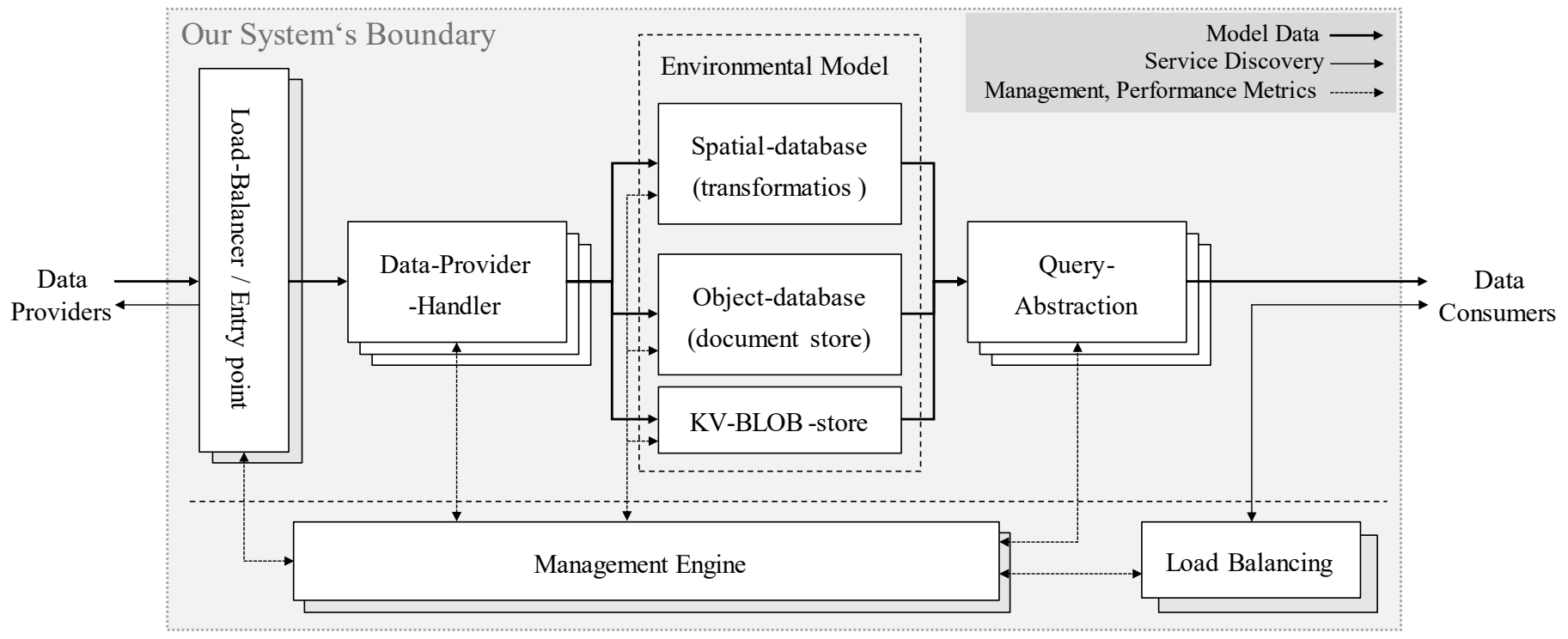

**Fig. 1.** Overview of the proposed software system. Modules in grey are slaves of the respective module, mirroring its state, for failover.

Targeting the systems robustness and to minimize the effect of crashed instances of software modules, all communication sessions held over the system's boundary shall be executed in their own process – thus not taking down any other instances of the same modules in the event of a crash, opposed to simple parallelizing by threading. All modules shall further be able to be run on separate machines, requiring not more than a network connection, thus enabling horizontal scaling in terms of computation power, memory and network link capacity, and allowing failover to different hardware. For modules only necessary once, in this case the "Load-Balancer/Entry point" (see **Fig. 1**), the "Management Engine" and the consumer side "Load Balancer", there shall be two instances running in master/slave-mode, effectively mirroring each other, on different machines. In the event of an instance failing or connection dropping, the second instance will take over operation, thus introducing a level of robustness and failure tolerance against network and machine failure.

The overall composition of the proposed software system can be found in **Fig. 1**. We will elaborate on the single modules in order of data-flow, from left to right.

**Load Balancer, Entry Point.**
The load balancer is the entry point from the outside to the system. It will request a new "Data Provider Handler" - on the machine currently assigned to spawn new instances, depending on the available machines loads - when a new data provider supplies data. It keeps track of the assignment of data providers to the "Data Provider Handlers" and routes incoming data to the appropriate handler. This internal routing allows the elimination of session handling in the data providers and provide a session- and state-less interface to the outside world, while still enabling internal caching and load balancing. The load-balancer in its function as entry-point to the system shall provide its address via broadcasts to ease data-provider hook-up, and allow system reconfiguration and migration without explicitly distributing changed addresses.

**Data Provider Handler, Lookup caching.**

The "Data Provider Handlers" in their simplest form relay the incoming data to the appropriate databases by performing the necessary queries. Their second purpose is ID-lookup and -transformation, speeding up this process by caching the required data. For illustration consider the following example:

A QR-code-marker has been detected by a 2D-camera. The "sensor-driver" (data provider) provides the following information about itself to the handler: ID and type, as well as information about the detected marker: type, ID (content) and the pose (transformation) in the camera's coordinate frame. Message (pseudocode):

```
{ [sensorID:foo, sensorType:camera],
  [type:marker.QR, markerID:bar, TF:<Mat>]  }
```

The problem is, that the sensor-driver does not provide the models internal ID of the marker, this has to be solved with an ID-lookup. The handler's function is to cache the internal IDs of all elements to be recognized by the sensor from the database, to allow fast ID-lookup and data relaying to the spatial database without expensive database queries. This keeps network traffic and computational load down, while also relieving the objects database server – at the cost of increased memory usage on the machines running the data provider handlers. A separate handler is instantiated for each data provider.

**Environmental Model.**

The spatial model will consist of three databases. One "spatial-database" building the graph of transformations like outlined in section 4.2. One document-store, the "Objects-database", enabling storage of (abstract) objects and assigning arbitrary attributes and corresponding values to them. Last, a separate key/value-store for persistent large binary objects (BLOB) like CAD-models, large images, etc. to take load from the objects data base itself and speed up querying.

The spatial database will ultimately be a custom design, for the remaining two existing technology may be used. Regarding the design of the database we will not elaborat in detail at this point. To satisfy the requirements regarding robustness and failure tolerance, the following requirements are posed on all databases: The databases shall support replication to different machines, automatic failover to different machines carrying copies of the database, and load balancing over their replicated instances or by database sharding. Aiming to support automatic and continuous change-feeds to consumers, we also require the objects- and spatial-database to support change-feeds natively. This eliminates the need for constant polling and querying of the databases with high frequency to supply up to date change-feeds, thus reducing the overall load on the system.

**Query-Abstraction.**

Query-Abstraction modules are instantiated per individual query to the system. They either process a single query for request/response-type queries to RAIL and shutdown afterwards, or keep running and keep a connection to the consumer requesting the query, in the case of a query having requested a change-feed. Their purpose is to first split the posed query into individual queries to the appropriate databases, and aggregating the databases responses in a single message to be delivered as response to the requesting customer.

**Load-Balancer, consumer-side.**
The load-balancer on the consumer-side does work differently from the load-balancer on the data-provider-side. Where the data is actually passed through the load-balancer on the data-provider side, the load-balancer on the consumer-side serves as a lookup-service, providing the address of the machine currently assigned to instantiate new query-abstraction modules. Similarly to the data-provider-side, the load-balancer shall broadcast its address to allow easy hook-up of consumers.

**Management Engine.**
The management engine constantly monitors state and availability of all modules, and spawns new modules in case the currently running ones crashed or become unavailable. Further it shall monitor performance metrics of the machines hosting the modules and select appropriate machines to instantiate new data-provider handler and query-abstraction modules on, communicating the new addresses to the load-balancers. The management-engine shall further serve as a central storage for configuration data used by the internal modules.

## 4 Conclusion

We have drafted the design of a software system capable of representing a dynamic industrial indoor environment, the goal of which is enabling LBS. In addition to the industrial application's requirements we have considered relevant quality measures of software systems such as scalability, robustness and performance and have designed a software architecture that fits these needs. The long term goal with this is to provide an open standard upon which LBS can be developed.

The software system and data modelling have been drafted to meet therequirements outlined in section. The software system addresses important considerations like robustness and scalability. Still, whether the system meets the requirements in production depends on additional factors like the used hardware and infrastructure. In summary, as open and flexible as the system is in its current definition, as much does performance depend on how it will be used. This includes questions like what data will be put into the model and how efficient the concrete algorithms will be. That is why guides for deployment and operations of the system will have to be developed alongside the system itself.

In further work we will proceed to implement the proposed concept to document its viability. This includes developing an accompanying toolset which aids tasks such as e.g. map creation.

## Acknowledgements

Research was funded under the project IIL by the European Fund for Regional Development (EFRE).

EUROPÄISCHE UNION
Europäischer Fonds für regionale Entwicklung

## References

1. Dhar, S., Varshney, U.: Challenges and business models for mobile location-based services and advertising Commun. ACM **54**(5), 121 (2011). doi: 10.1145/1941487.1941515
2. Xu, Y., Chen, X., Ma, L.: LBS based disaster and emergency management. In: Liu, Y. (ed) 18th International Conference on Geoinformatics, 2010: 18 - 20 June 2010, Beijing, China. 2010 18th International Conference on Geoinformatics, Beijing, China, pp. 1–5. IEEE, Piscataway, NJ (2010). doi: 10.1109/GEOINFORMATICS.2010.5567872
3. Fritsch, L., Scherner, T.: A Multilaterally Secure, Privacy-Friendly Location-Based Service for Disaster Management and Civil Protection. In: Lorenz, P., Dini, P. (eds) Networking - ICN 2005: 4th International Conference on Networking, Reunion Island, France, April 17-21, 2005, Proceedings, Part II. Lecture Notes in Computer Science, vol. 3421, pp. 1130–1137. Springer-Verlag Berlin Heidelberg, Berlin, Heidelberg (2005)
4. Google Public Alerts, https://developers.google.com/public-alerts/, last accessed 2017/11/03.
5. Barnes: Location-Based Services: The State of the Art e-Service Journal **2**(3), 59 (2003). doi: 10.2979/esj.2003.2.3.59
6. Infsoft Whitepaper, https://www.infsoft.de/portals/0/images/solutions/basics/whitepaper/infsoft-whitepaper-de-indoor-positionsbestimmung_download.pdf, last accessed 2017/11/03.
7. IndoorLBS, http://www.indoorlbs.com/, last accessed 2017/11/03.
8. IndoorAtlas, http://www.indooratlas.com/, last accessed 2017/11/03.
9. estimote, https://estimote.com/, last accessed 2017/11/03.
10. Biczok, G., Diez Martinez, S., Jelle, T., Krogstie, J.: Navigating MazeMap: Indoor human mobility, spatio-logical ties and future potential. In: 2014 IEEE International Conference on Pervasive Computing and Communication Workshops (PERCOM WORKSHOPS), pp. 266–271 (2014). doi: 10.1109/PerComW.2014.6815215
11. micello, https://www.micello.com/, last accessed 2017/11/03.
12. Ubisense SmartFactory, https://ubisense.net/de/products/smart-factory, last accessed 2017/11/03.
13. IntraNav, http://intranav.com/, last accessed 2017/11/03.
14. Gunduz, M., Isikdag, U., Basaraner, M.: A REVIEW OF RECENT RESEARCH IN INDOOR MODELLING & MAPPING Int. Arch. Photogramm. Remote Sens. Spatial Inf. Sci. **XLI-B4**, 289–294 (2016). doi: 10.5194/isprsarchives-XLI-B4-289-2016
15. Afyouni, I., Ray, C., Claramunt, C.: Spatial models for context-aware indoor navigation systems: A survey JOSIS(4) (2012). doi: 10.5311/JOSIS.2012.4.73

16. Foote, T.: tf: The transform library. In: 2013 IEEE International Conference on Technologies for Practical Robot Applications (TePRA): 22 - 23 April 2013, Woburn, Massachusetts, USA. 2013 IEEE Conference on Technologies for Practical Robot Applications (TePRA), Woburn, MA, USA, pp. 1–6. IEEE, Piscataway, NJ (2013). doi: 10.1109/TePRA.2013.6556373
17. Zlatanova, S., Sithole, G., Nakagawa, M., Zhu, Q.: Problems in indoor mapping and modelling Acquisition and Modelling of Indoor and Enclosed Environments 2013, Cape Town, South Africa, 11-13 December 2013, ISPRS Archives Volume XL-4/W4, 2013 (2013)
18. CityGML: Exchange and Storage of Virtual 3D City Models, http://www.citygml.de, last accessed 2017/11/03.
19. GeoJSON, http://geojson.org/, last accessed 2017/11/03.